\documentclass[lettersize,journal]{IEEEtran}
\usepackage{amsmath,amsfonts}
\usepackage{amssymb}
\usepackage{algorithmic}
\usepackage{algorithm}
\usepackage{array}
\usepackage[caption=false,font=normalsize,labelfont=sf,textfont=sf]{subfig}
\usepackage{textcomp}
\usepackage{stfloats}
\usepackage{tabularx}
\usepackage{url}
\usepackage{makecell}
\usepackage{verbatim}
\usepackage{graphicx}
\usepackage{cite}
\usepackage[percent]{overpic}

\usepackage{adjustbox}
\usepackage{booktabs}
\usepackage[table]{xcolor}
\usepackage{svg} 
\usepackage{multirow} 
\usepackage{amssymb}
\usepackage{pifont}
\newcommand{\cmark}{\textcolor{green!60!black}{\ding{51}}}
\newcommand{\xmark}{\textcolor{red}{\ding{55}}}

\newcommand\normx[1]{\left\Vert#1\right\Vert}

\makeatletter
\newcommand\notsotiny{\@setfontsize\notsotiny\@vipt\@viipt}
\makeatother

\usepackage{tikz}
\usepackage{xcolor}
\usepackage{hyperref}

\definecolor{deepgreen}{RGB}{0,100,0}
\definecolor{darkred}{RGB}{140, 30, 30}
\hypersetup{
    colorlinks=true,
    citecolor=deepgreen,
    linkcolor=darkred,
    urlcolor=blue
}

\newcommand{\fourdiagcell}{%
\begin{tikzpicture}[
    baseline=(current bounding box.center),
    every node/.style={
        font=\footnotesize\bfseries,
        inner sep=0pt
    }
]
    \def\W{1.95}
    \def\H{0.92}

    \path[use as bounding box] (0,0) rectangle (\W,\H);

    \draw[line width=0.25pt] (0,\H) -- (0.60*\W,0);

    \node at (1.8,0.830) {Dataset};
    \node at (1.8,0.35) {Seq.};
    \node at (1.80,0.00) {Len.(m)};

    \node at (0.45,0.0) {Method};

\end{tikzpicture}%
}

\usepackage{caption}

\DeclareMathSizes{10}{9}{7}{5}

\title{\LARGE \bf Does Robust VIO Need More Learning? \\Geometry-Verified Visual Measurements under Distribution Shift}

\author{Yangyang Ning, Shu Liang, Quanbo Ge, 
Tianchen Deng, Yuhua Qi, and Shenghai Yuan%
\thanks{Y. Ning and S. Liang are with the Department of Control Science and Engineering, Tongji University, Shanghai 201804, China.
Q. Ge is with the School of Automation, Nanjing University of Information Science and Technology, Nanjing, Jiangsu 210044, China.
T. Deng is with Shanghai Jiao Tong University, Shanghai, China.
Y. Qi is with Sun Yat-sen University, China.
S. Yuan is with Nanyang Technological University, Singapore.}}

\begin{document}

\bstctlcite{IEEEexample:BSTcontrol}

\maketitle
\thispagestyle{empty}
\pagestyle{empty}

\begin{abstract}
Learning is increasingly introduced into visual--inertial odometry (VIO), ranging from learned feature front-ends to learning-dominant motion and geometry estimation. However, learning more of the pipeline does not necessarily improve robustness when deployment conditions differ from the training distribution. This work asks whether robust VIO under distribution shift truly requires deeper learned estimation, or whether learning can be confined to visual measurement generation. We propose a minimal-learning stereo VIO framework in which SEA-RAFT is used only to propose dense stereo correspondences and predict their uncertainty, while temporal tracking, geometric verification, and state estimation remain explicit. Dense flow is sampled at sparse feature locations, filtered using predicted uncertainty and stereo epipolar consistency, and incorporated into a sliding-window stereo-inertial estimator through uncertainty-weighted reprojection factors. The same uncertainty is further propagated through stereo triangulation for downstream anisotropic 3D Gaussian mapping. Experiments on EuRoC, VIODE, and 4Seasons demonstrate accurate and stable estimation under motion blur, dynamic scenes, illumination changes, and large indoor-to-outdoor distribution shifts. Ablations show that learned flow alone is insufficient: the gains arise from combining learned correspondence proposals with geometric verification and uncertainty-aware weighting. These results suggest that, for OOD-robust VIO, carefully integrated learned visual measurements can be more effective than learning a larger fraction of the estimation pipeline. Code and configs for the benchmark will be open-source upon acceptance.  A supplementary video is available \href{https://drive.google.com/file/d/1EVRhOkhanmNXHbQS1Vr80FoEIAYOYOV2/view?usp=sharing}{here}.
\end{abstract}


\section{Introduction}\label{sec1}
Visual--inertial odometry (VIO) is essential for various robotic tasks, such as manipulation and navigation \cite{RN309,RN91,RN398}. Classical VIO represents visual measurements through explicit geometric residuals, making the estimation process efficient, interpretable, and physically verifiable \cite{RN809,RN541,RN298,RN517,RN586}. However, classical hand-crafted features can degrade under motion blur, weak texture, illumination changes, repetitive structures, fast motion, and occlusion \cite{RN303,RN302,RN304}. Learning has therefore been increasingly introduced into the modern VIO pipeline. 
This raises a central question: \textbf{Does modern VIO really require learning a larger fraction of the estimation pipeline when facing distribution shifts?}

\begin{figure}[t]
\centering
\begin{overpic}[width=0.95\linewidth]{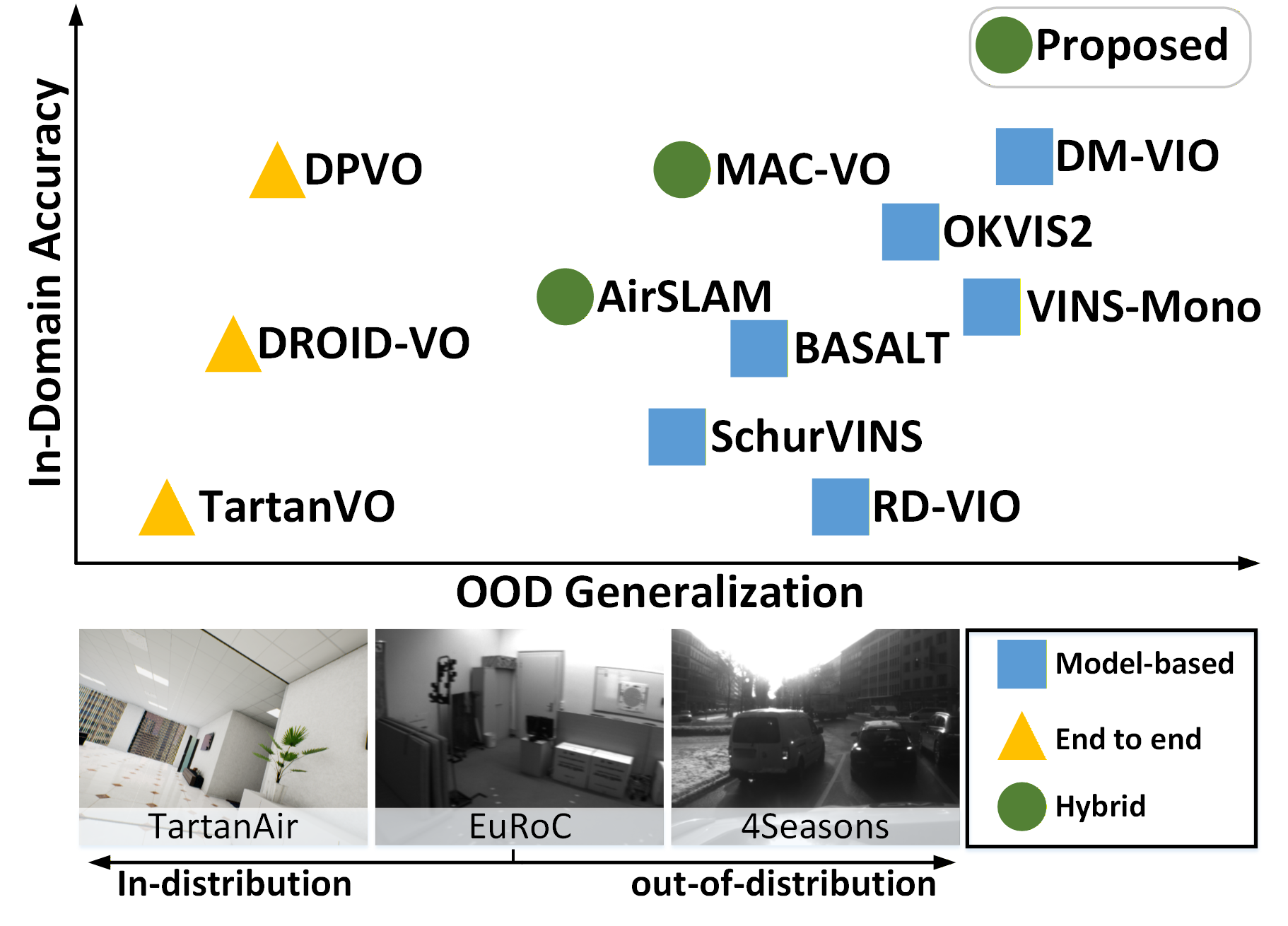}

\put(33,58.0){\scriptsize\cite{teed2023deep}}       
\put(36,44.5){\scriptsize\cite{teed2021droid}}      
\put(30,32){\scriptsize\cite{wang2021tartanvo}}   

\put(59,48){\scriptsize\cite{RN1293}}             
\put(69,58){\scriptsize\cite{qiu2025mac}}         

\put(73,43.8){\scriptsize\cite{RN474}}   
\put(71,38){\scriptsize\cite{RN1291}}  
\put(79,32){\scriptsize\cite{RN1279}}        
\put(84,53){\scriptsize\cite{leutenegger2022okvis2}} 
\put(94,59){\scriptsize\cite{RN455}}  
\put(97,47){\scriptsize\cite{RN91}}        

\end{overpic}

\caption{Qualitative positioning of representative geometric,
learning-dominant, and hybrid VO/VIO methods in terms of in-domain
accuracy and OOD robustness. The placement is based on the experimental
trends observed across EuRoC, VIODE, and 4Seasons.}
\label{fig:motivation}
\vspace{-3mm}
\end{figure}
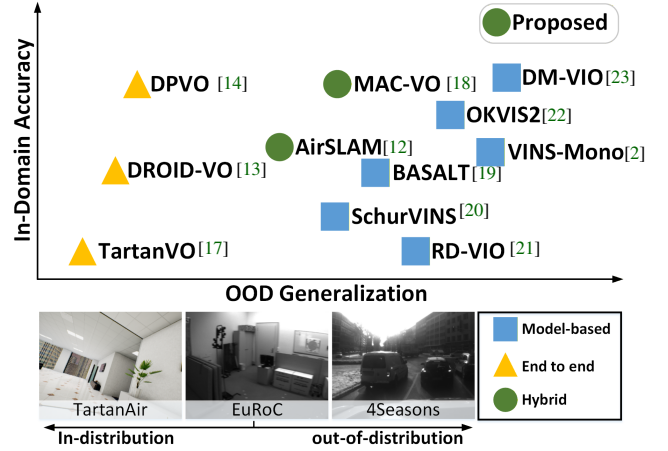

Existing learning-based VO/VIO systems cover a broad spectrum. At the
lightweight end, learned local features and matchers, such as SuperPoint,
LightGlue, and OmniGlue, replace hand-crafted feature detection and association
while leaving the remaining geometric pipeline unchanged. AirSLAM \cite{RN1293} further
introduces learned point and line features through PLNet while retaining a
conventional optimization-based backend. At the other end,
learning-dominant systems, including DROID-SLAM, DPVO, MASt3R-SLAM, and
VGGT-based SLAM, embed learned correspondence, depth, motion, pose, or scene
reconstruction more deeply into the estimation process
\cite{teed2021droid,teed2023deep,murai2025mast3r,wang2025vggt}.
Differentiable geometry and end-to-end odometry networks further increase the
proportion of the pipeline governed by learned representations. Despite their strong performance on standard benchmarks,
all learned components remain dependent on the distributions represented
during training \cite{wang2021tartanvo}. In practical deployment, the
testing environments rarely match the training data exactly. Changes in scene
appearance, motion patterns, illumination, dynamic scene, sensor characteristics, and
environmental structure introduce distribution shifts, placing these systems
in out-of-distribution (OOD) conditions where their learned predictions may
become unreliable (see Fig. \ref{fig:motivation}).


The key challenge is therefore not simply to introduce a deeper or more expressive
network. Increasing model capacity may improve in-distribution accuracy, but it can
also strengthen dependence on the training distribution and make failure modes more
difficult to interpret. What is required is an architecture that retains the
representational advantages of learning while preserving the interpretability and
stability of geometric estimation under OOD conditions. Learned visual priors can
provide richer correspondence and motion cues
\cite{teed2020raft,wang2024sea,kang2024visual,kang2025deepof}, but their outputs
should not be inserted into a VIO estimator as deterministic measurements. In
particular, network-predicted confidence or uncertainty is not automatically the
covariance required by a geometric residual \cite{RN470,RN533,RN1156}. Without
estimator-compatible uncertainty and explicit geometric verification, learned
measurements may introduce overconfident or inconsistent constraints, causing the
estimator to become unstable precisely when the system encounters distribution shift
\cite{RN1011,RN1199,RN568}.

To address this challenge, we propose a \textbf{geometry-verified,
uncertainty-aware stereo VIO framework}. SEA-RAFT is used only to generate
dense stereo correspondences and their uncertainty
\cite{teed2020raft,wang2024sea,kang2025deepof}. The correspondences are
verified by learned uncertainty filtering and geometric stereo epipolar consistency. Their predicted uncertainty is then converted into
covariance-weighted reprojection factors in a sliding-window VIO backend
\cite{RN400}. Thus, learning proposes visual measurements, geometry
verifies them, and covariance controls their influence on state estimation.
The same uncertainty is propagated through stereo triangulation to 3D Gaussian mapping.

The contributions are summarized as follows:
\begin{itemize}
    \item We formulate OOD-robust stereo VIO as a \textbf{learned visual
    measurement modeling} problem, while keeping geometric state estimation
    explicit.

    \item We propose a lightweight visual front-end that combines SEA-RAFT
    correspondence and uncertainty with stereo geometric
    verification.

    \item We incorporate the verified observations as covariance-weighted VIO
    factors and propagate their uncertainty to 3D Gaussian mapping. Experiments
    show stronger robustness than the compared geometric, hybrid, and
    learning-dominant methods under distribution shift.
\end{itemize}

\section{Related Work}

\noindent \textbf{Geometric VIO and Lightweight Learning.}
Classical VIO systems \cite{RN517,RN586}, such as OKVIS, VINS-Mono, DM-VIO, and SchurVINS, formulate visual measurements as explicit reprojection factors in filtering or sliding-window optimization backends
\cite{RN309,RN91,RN455,RN1291}. This structure provides computational efficiency, geometric interpretability, and stable state estimation, but its performance remains limited by the robustness of hand-crafted feature detection, tracking, and association
\cite{RN303,RN302,RN809,RN541}. Lightweight learning-enhanced systems replace selected visual front-end components while retaining the geometric estimator. Learned features and matchers, such as SuperPoint, LightGlue, and OmniGlue, improve feature extraction and correspondence, while AirSLAM further introduces learned point and line features through PLNet
\cite{RN1293}. Although these methods reduce the dependence on hand-crafted features, their learned outputs are mainly treated as feature locations, descriptors, or associations. Their confidence is not converted into estimator-compatible measurement covariance, and geometrically inconsistent predictions may introduce unreliable constraints under distribution shift.

\noindent \textbf{Learning-Dominant Visual Geometry.}
At the other end of the spectrum, learning-dominant systems replace a larger part of the geometric pipeline with learned correspondence, motion, depth, pose, or scene representations. DROID-SLAM and DPVO use learned update operators for dense correspondence and iterative pose estimation, while MASt3R-SLAM and VGGT-based systems exploit large-scale visual priors for camera pose and 3D reconstruction
\cite{teed2021droid,teed2023deep,murai2025mast3r,wang2025vggt}. Differentiable geometry and end-to-end odometry networks further increase the proportion of the pipeline governed by learning. These methods can achieve strong accuracy on standard benchmarks, but their predictions remain dependent on the distributions represented during training. When scene appearance, motion, illumination, texture, or sensor characteristics change, the learned representation may become unreliable
\cite{wang2021tartanvo}. Their outputs are also not always designed as real-time VIO factors with explicit measurement covariance. In contrast, our method does not replace the geometric VIO estimator with a learned odometry model. It restricts learning to dense visual correspondence and studies how these predictions can be geometrically verified, converted into estimator-compatible uncertainty weights, and safely incorporated into a conventional VIO backend.

\noindent \textbf{Uncertainty-Aware Visual Measurement Modeling.}
Robust visual estimation has been studied through robust loss functions, graduated non-convexity, feature selection, and uncertainty-aware residual weighting
\cite{RN1011,RN1199,RN1243,RN1242,RN484}. Recent methods further use learned uncertainty or dense optical flow to improve VO/VIO robustness
\cite{RN470,teed2020raft,wang2024sea,kang2024visual,kang2025deepof}. However, the predicted uncertainty is commonly used for correspondence rejection, heuristic residual weighting, or direct pose optimization. Network confidence is not automatically equivalent to the covariance required by a geometric reprojection factor
\cite{RN533,RN1156}. Moreover, few methods explicitly combine learned uncertainty with stereo epipolar consistency and temporal reprojection consistency to determine whether a correspondence remains geometrically reliable under distribution shift. Our method addresses this problem by converting network-predicted flow uncertainty and geometric consistency into correspondence-level covariance, which is directly used in the VIO reprojection factors.

\noindent \textbf{Uncertainty-Aware Mapping.}
Recent neural and Gaussian-based mapping methods achieve strong reconstruction and rendering quality using dense visual observations, estimated depth, and learned geometric priors
\cite{RN1252,RN469,murai2025mast3r,kerbl20233dgaussian}. However, most methods focus on map appearance or reconstruction accuracy, while the uncertainty of the upstream visual measurements is not explicitly propagated into the map representation. This uncertainty is important in VIO-based mapping because inaccurate visual constraints affect both pose estimation and 3D structure initialization
\cite{RN309,RN474,RN531,RN766}. Our method propagates the uncertainty-informed visual measurement covariance through stereo triangulation to 3D Gaussian mapping, enabling uncertainty-aware Gaussian initialization and confidence weighting.

\noindent\textbf{Positioning of This Work.}
Existing learning-enhanced VO/VIO methods either use learning as a stronger visual front-end \cite{RN1293} or rely on learned geometric priors to estimate correspondence, motion, depth, or 3D structure \cite{teed2021droid,teed2023deep,murai2025mast3r,wang2025vggt}.
\textit{In contrast, we ask whether more learning is always beneficial under distribution shift.}
We follow a lightweight learning route: learning generates dense correspondence and uncertainty \cite{teed2020raft,wang2024sea,kang2024visual,kang2025deepof}, while geometric verification and estimator-compatible covariance determine how each observation enters the VIO estimator \cite{RN470,RN533,RN1156}.
Specifically, we treat learned optical flow as a stochastic visual observation, construct an uncertainty-informed covariance proxy using network uncertainty and stereo epipolar consistency, then use it in VIO factors and propagate it to 3D Gaussian mapping. Experiments under visual degradation and distribution shift show improved VIO robustness without replacing the geometric estimator or using deeper front-end features.

\section{Methodology}
\subsection{Problem Formulation and System Overview}






In VIO, the optimization problem is constrained by both inertial and visual residuals. The effectiveness of these constraints depends on two critical factors: quality (accuracy required to keep the estimator unbiased) and quantity (abundance required to ensure resilience against noise and disturbances). Inertial residuals are formed by pre-integration between image frames and are inherently unaffected by visual appearance changes. 
Visual residuals, on the other hand, are highly susceptible to degraded environments like motion blur, illumination changes, or textureless scenes. When handling these conditions, existing methods expose a strict trade-off between tracking quality and quantity. Traditional algorithms (e.g., KLT, ORB, SIFT) yield high-quality but low-quantity tracks under stress, leaving the system vulnerable to noise. Conversely, end-to-end learning trackers provide high-quantity but low-quality constraints, robustly surviving extreme conditions but suffering from large errors in OOD scenarios.

To bridge this gap, this work proposes a hybrid frontend that combines a traditional tracker with a geometrically verified learning-based tracker. This design improves the reliability of visual factors while retaining sufficient visual constraints for state estimation. It ensures robustness in extreme visual conditions while mitigating performance degradation in OOD scenarios. Furthermore, we leverage learned optical flow uncertainty and stereo geometric constraints (epipolar and reprojection consistency) to propagate a dense ellipsoidal map, while also formulating uncertainty-aware visual factors for backend optimization.

\begin{figure*}[t]
\centering
\includegraphics[width=0.9\linewidth]{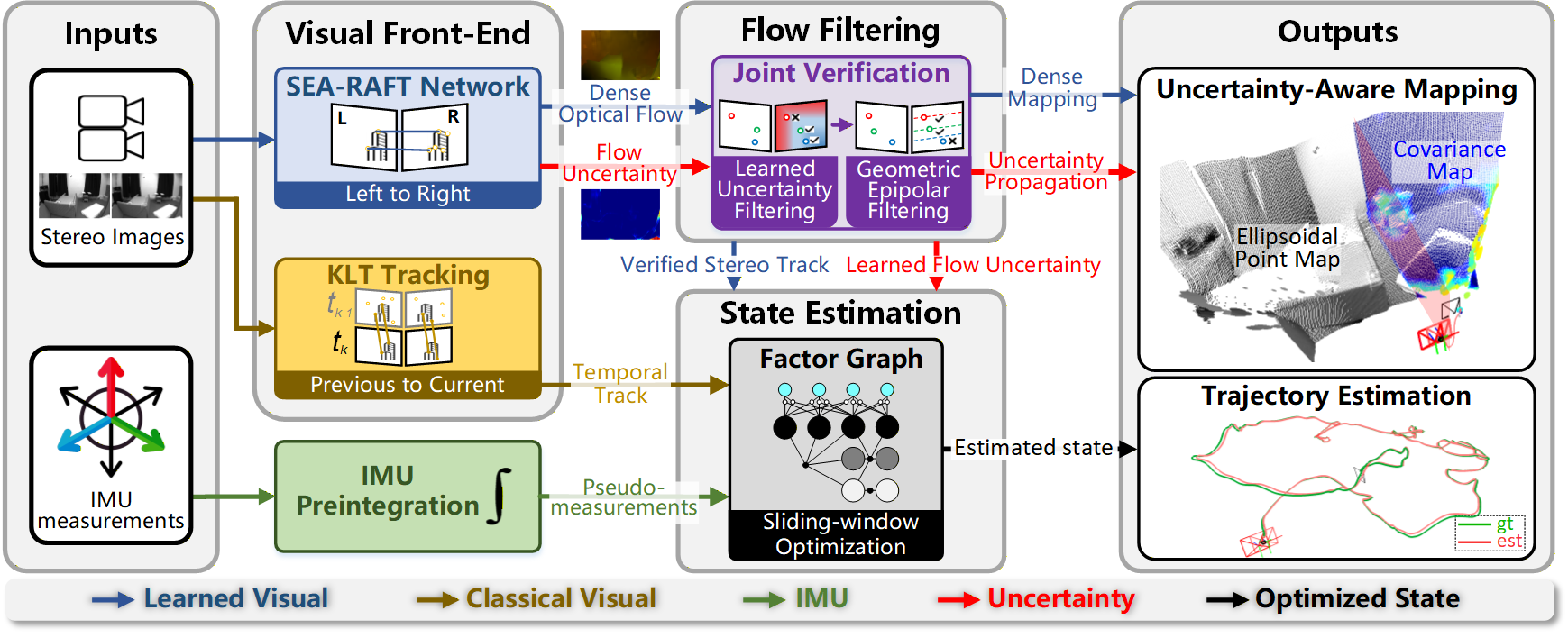}
\vspace{-5pt}
\caption{Overview of the proposed hybrid system pipeline. Geometric-verified learned optical flow is utilized for stereo tracking, while the learned uncertainty is leveraged for visual factor weighting and ellipsoidal map reconstruction.}
\label{fig:system}
\vspace{-8pt}
\end{figure*}

Given a stereo image sequence $\{I_l^k, I_r^k\}_{k=1}^{N}$ and the collection of IMU measurements between consecutive frames $\{\mathcal{U}_{k,k+1}\}_{k=1}^{N-1}$, stereo VIO estimates the robot states
\[
\mathcal{X}=\{\mathbf{R}_k,\mathbf{p}_k,\mathbf{v}_k,\mathbf{b}_{g,k},\mathbf{b}_{a,k}\}_{k=1}^{N},
\]
where $\mathbf{R}_k$, $\mathbf{p}_k$, $\mathbf{v}_k$, $\mathbf{b}_{g,k}$, and $\mathbf{b}_{a,k}$ denote orientation, position, velocity, gyroscope bias, and accelerometer bias, respectively.
In classical VIO, each visual observation is represented by a reprojection residual with a predefined image-domain covariance. In contrast, learned optical flow produces dense visual predictions whose uncertainty is data-driven and not directly equivalent to the covariance required by the VIO estimator. Therefore, the main problem considered in this work is how to convert learned dense visual observations into estimator-compatible probabilistic measurements.


Visual observations are modeled using sparse optical flow, integrating both conventional temporal tracking and learned stereo matching. The feature tracking pipeline executes in three sequential steps: temporal tracking, feature detection, and stereo tracking.

First, temporal tracking is performed across consecutive frames. Existing features are tracked independently within the respective left and right image streams (i.e., $I_l^{k-1} \rightarrow I_l^k$ and $I_r^{k-1} \rightarrow I_r^k$) using conventional Kanade-Lucas-Tomasi (KLT) flow with predefined covariance $\boldsymbol{\Sigma}_{\text{obs}}$:
\begin{equation}
\begin{aligned}
    \mathbf{Z}_{\text{cur}}&\sim\mathcal{N}\left(\mathbf{z}_{\text{pre}} + \mathbf{f}_{\text{klt}},\boldsymbol{\Sigma}_{\text{obs}}\right),\\
    \mathbf{Z}_{\text{pre}}&\sim\mathcal{N}\left({\mathbf{z}}_{\text{pre}}, \boldsymbol{\Sigma}_{\text{obs}}\right),
\end{aligned}
\end{equation}
where $\mathbf{z}_{\text{pre}}$ is the previously tracked or detected feature, $\mathbf{f}_{\text{klt}}$ is the KLT flow vector, and $\mathbf{Z}_{\text{cur}},\mathbf{Z}_{\text{pre}}$ is the updated feature distribution in the current and previous frame. Then, feature detection is applied. New FAST features, denoted as $\mathbf{z}_l$, are extracted in the current left frame to replenish the feature pool with predefined covariance:
\begin{equation}
     \mathbf{Z}_l\sim\mathcal{N}\left({\mathbf{z}}_l, \boldsymbol{\Sigma}_{\text{obs}}\right),
\end{equation}
Finally, stereo tracking maps these newly detected features from the left image to the right image using a learned optical flow network with propagated learning covariance $\boldsymbol{\Sigma}_{\text{pro}}$:
\begin{equation}
     \mathbf{Z}_r \sim\mathcal{N}\left( \mathbf{z}_l + \mathbf{f}_{\text{net}}, \boldsymbol{\Sigma}_{\text{pro}}\right),
\end{equation}
where $\mathbf{f}_{\text{net}}$ is the learned stereo flow and $\mathbf{Z}_r$ is the corresponding tracked feature distribution in the right frame.

These 2D feature covariances are then propagated downstream, serving as the uncertainty source for 3D Gaussian mapping and acting as uncertainty weights in the visual-inertial bundle adjustment (see Fig.~\ref{fig:system} for system overview).

\subsection{Hybrid Learned-Geometric Visual Observation Generation}

Conventional Kanade-Lucas-Tomasi (KLT) tracking is widely used in VIO for its real-time efficiency and sub-pixel accuracy. However, its strict reliance on local brightness constancy makes it highly vulnerable to visual degradations like motion blur, weak textures, and sudden illumination changes. In essence, KLT is estimator-friendly but visually fragile.

Conversely, learning-based optical flow methods (e.g., SEA-RAFT) leverage global context to maintain robust dense correspondences even under severe degradation. Yet, these raw dense outputs lack multi-view geometric rigor. They are prone to producing confident but false matches over extreme visual conditions or in OOD scenes, making them ideal correspondence proposers but unreliable standalone measurements.

To bridge this gap, we adopt a hybrid strategy pairing an efficient conventional temporal tracker with a robust learning-based stereo tracker named SEA-RAFT \cite{wang2024sea}. The stereo module utilizes a geometry-aided dense-to-sparse extraction technique, bilinearly sampling the network's dense flow at sparse sub-pixel locations and jointly verifying them via epipolar constraints and network-predicted uncertainty. 

The input to SEA-RAFT consists of two three-channel RGB raw images $I_l, I_r\in\mathbb{R}^{H\times W\times3}$, and the output is a tensor in $\mathbb{R}^{H\times W\times6}$. The first two channels represent the displacement components of the dense optical flow along the $u$ and $v$ image axes, denoted as $\boldsymbol{\mu}\in\mathbb{R}^{H\times W\times2}$. For a set of newly detected FAST features $\{\mathbf{z}^{1}_{l},\mathbf{z}^{2}_{l},\dots,\mathbf{z}^{k}_l\}$ in the left image $I_l$, their corresponding feature locations $ \mathbf{z}_r$ in the right image are determined by bilinearly interpolating the dense flow map at the feature coordinates. For a point $\mathbf{z}^{i}_{l}$, let $\mathcal{N}(\mathbf{z}^{i}_{l})$ be the set of its four nearest discrete neighbors. The corresponding right feature is calculated by:
\begin{equation}
   \mathbf{z}^{i}_{r} = \mathbf{z}^{i}_{l} + \sum_{\mathbf{x} \in \mathcal{N}(\mathbf{z}^{i}_{l})} w_{\mathbf{x}} \boldsymbol{\mu}(\mathbf{x}),
\end{equation}
where $w_{\mathbf{x}}$ represents the bilinear interpolation weight corresponding to the distance between $\mathbf{z}_l^i$ and grid node $\mathbf{x}$.

\begin{figure}[t]
\centering
\includegraphics[width=0.95\linewidth]{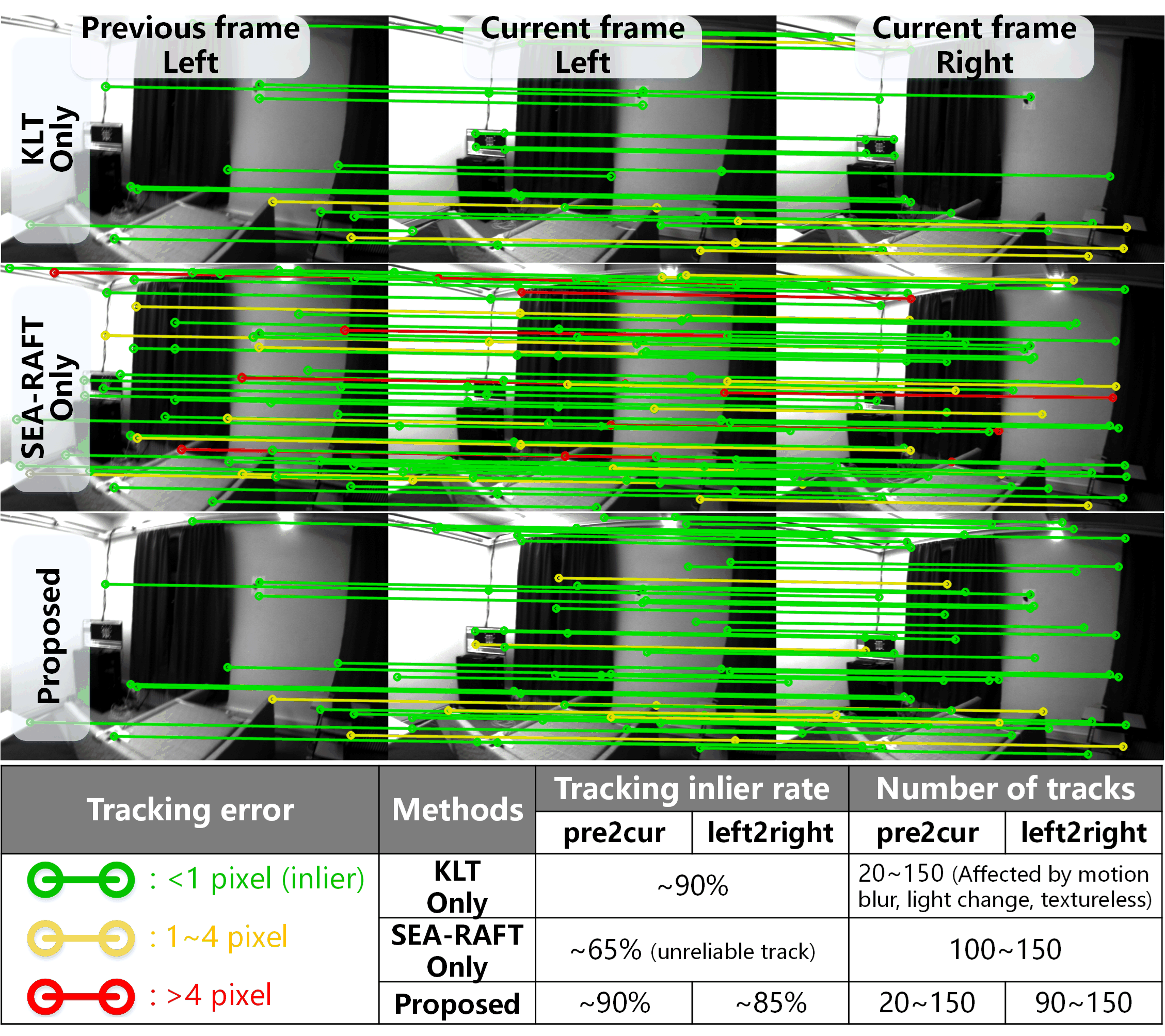}
\caption{Qualitative and quantitative comparison of temporal tracking and stereo matching on EuRoC-style sequences. KLT-only tracking is only able to \textbf{track sparsely at corners}, while prone to loss in extreme visual conditions and featureless areas. SEA-RAFT-only matching provides denser correspondences but still introduces \textbf{geometrically inconsistent tracks}. The proposed uncertainty-aware selection preserves dense and consistent correspondences across temporal and stereo frames. This also shows that data-driven methods alone can still fail under OOD conditions.}
\label{fig:matching}
\vspace{-3mm}
\end{figure}

The dense-to-sparse feature tracks are then jointly verified by learned flow uncertainty and epipolar constraints. The learned flow uncertainty comes from the last four channels of SEA-RAFT output, which contain weighting coefficients for normal conditions, $\boldsymbol{\alpha}_1,\boldsymbol{\alpha}_2\in\mathbb{R}^{H\times W}$, and weighting coefficients for anomalous occlusion cases, $\boldsymbol{\beta}_1,\boldsymbol{\beta}_2\in\mathbb{R}^{H\times W}$. 
Following the official SEA-RAFT implementation \cite{wang2024sea}, which makes the mixture logits and log-scale channels explicit, the appropriate flow uncertainty distribution $\sigma\in\mathbb{R}^{H\times W}$ is calculated by the mixture Laplacian loss definition as follows:
\begin{equation}
\boldsymbol{\sigma} = \frac{\exp{(\boldsymbol{\alpha}_1)}}{\exp{(\boldsymbol{\alpha}_1)} + \exp{(\boldsymbol{\alpha}_2)}} \boldsymbol{\beta}_1 + \frac{\exp{(\boldsymbol{\alpha}_2)}}{\exp{(\boldsymbol{\alpha}_1)} + \exp{(\boldsymbol{\alpha}_2)}} \boldsymbol{\beta}_2.
\end{equation}
By normalizing this uncertainty distribution, we obtain a range of $[0,1]$ for the normalized flow uncertainty $\hat{\boldsymbol{\sigma}}\in\mathbb{R}^{H\times W}$:
\begin{equation}
\hat{\boldsymbol{\sigma}} = \frac{\boldsymbol{\sigma} - \min(\boldsymbol{\sigma})}{\max(\boldsymbol{\sigma}) - \min(\boldsymbol{\sigma}) +\epsilon },
\end{equation}
where $\epsilon$ is for numerical stability.
To eliminate anomalous occlusions, we first discard feature tracks with high learned flow uncertainty, retaining only those that satisfy $\hat{\boldsymbol{\sigma}}(\mathbf{z}^{i}_{l}) < \epsilon_{\text{net}}$ where $\hat{\boldsymbol{\sigma}}(\mathbf{z}^{i}_{l})$ denotes the bilinear interpolated uncertainty at feature pixel location $\mathbf{z}^{i}_{l}$. These uncertainty-verified tracks are then further filtered using geometric epipolar constraints:
\begin{equation}
\left|\hat{\mathbf{z}}_l^{\intercal}\;\mathbf{E}_{I_l,I_r}\;\hat{\mathbf{z}}_r\right|\leq\epsilon_{\operatorname{geo}},
\end{equation}
where $\hat{\mathbf{z}}_l,\hat{\mathbf{z}}_r\in\mathbb{R}^3$ are stereo-tracked feature unit direction vectors from the left and right camera frames via back-projection, $\mathbf{E}_{I_l,I_r}$ is the essential matrix from calibrated stereo extrinsic. This joint verification process maximizes the quality of the dense-to-sparse feature tracks without sacrificing their quantity, ensuring robust and plentiful visual constraints for backend optimization (see Fig.~\ref{fig:matching}).

\subsection{Uncertainty-Aware Stereo VIO}

In conventional VIO, visual reprojection residuals are weighted by the inverse covariance of predefined Gaussian image noise, $\boldsymbol{\Sigma}_{\mathrm{obs}}$. In this work, the uncertainty predicted by the stereo flow network is used as a relative reliability measure to adaptively weight learned correspondences. Assuming that the predefined observation noise and the flow-induced uncertainty are independent, the covariance associated with the learned right-image observation is defined as
\begin{equation}
    \boldsymbol{\Sigma}_{\mathrm{pro}}
    =
    \boldsymbol{\Sigma}_{\mathrm{obs}}
    +
    \boldsymbol{\Sigma}_{\mathrm{net}},
\end{equation}
where $\boldsymbol{\Sigma}_{\mathrm{net}}
=
\lambda\hat{\sigma}^{2}(\mathbf{z}^{i}_{l})\mathbf{I}$
is an uncertainty-informed covariance proxy, and $\lambda$ is a fixed covariance scale in pixels squared. The normalized network uncertainty is not treated as a statistically calibrated image-domain covariance; instead, it provides relative adaptive weighting for the learned observations. 
The resulting covariance is then used to weight the reprojection residual associated with the learned flow observation:
\begin{equation}
    \begin{aligned}
        e_{\text{vis}}=\frac{1}{2}\normx{ \boldsymbol{\Sigma}_{\text{pro}}^{-\frac{1}{2}}
        \bigg(\boldsymbol{\pi}\left(\mathbf{T}_{I_{t}I_{h}}\mathbf{p}_l^h\right)-\mathbf{z}^{t}_r\bigg)
        }^2_2,
    \end{aligned}\label{e-vis}
\end{equation}
where $\boldsymbol{\pi}(\cdot)$ is the projection function, $\mathbf{T}_{I_{t}I_{h}}$ is the transformation mapping the host frame to the target frame, $\mathbf{p}^h_l$ is the 3D point coordinates in the host (left) frame, and $\mathbf{z}^{t}_r$ represents the 2D observation in the target (learned right) frame. Finally, this learned visual factor is incorporated into the sliding window optimization, where the optimal state is obtained via a \textit{maximum a posteriori} (MAP) estimation:
\begin{equation}
\begin{aligned}
    \mathcal{X}^*:=&\text{arg }\underset{\mathcal{X}}{\text{min}}\;\Bigg\{  \sum_{\mathbf{x}\in\mathcal{M}}\frac{1}{2}\normx{\mathbf{r}_m+\mathbf{J}_m\left(\mathbf{x}-\Bar{\mathbf{x}}\right)}_2^2 +   \\
    &\sum_{\mathbf{x}\in\mathcal{F}}e_{\text{imu}}(\mathbf{x}) + \sum_{\mathbf{x}\in\{\mathcal{K},\mathcal{F}\},\;\mathbf{p}\in\mathcal{L}}\rho\left(e_{\text{vis}}(\mathbf{x},\mathbf{p})\right)\Bigg\},
\end{aligned}
\end{equation}
where the first term denotes the marginalization prior $\{\mathbf{J}_m,\mathbf{r}_m\}$ on the Markov blanket $\mathcal{M}$, $\Bar{\mathbf{x}}$ is the fixed linearization point from first-estimates Jacobians\cite{RN525}. The second term denotes the IMU preintegration error $e_{\text{imu}}(\cdot)$ onto the consecutive recent frames $\mathcal{F}$, and the last term denotes the visual reprojection error $e_{\text{vis}}(\cdot)$  onto consecutive recent frames and keyframes $\mathcal{K}$, $\rho(\cdot)$ denotes the robust loss function. 

In conclusion, the learned weights act as an adaptive constraint during the geometric-inertial optimization, effectively mitigating the impact of uncertain flows caused by occlusions to preserve global state consistency.

\subsection{Uncertainty Propagation to 3D Gaussian Mapping}

Beyond weighting the VIO reprojection factors, the estimator-compatible 2D visual covariance can also be propagated through stereo triangulation to formulate 3D landmark covariances, thereby facilitating 3D Gaussian mapping. Specifically, we leverage the SEA-RAFT flow uncertainty as the 2D observation noise. By propagating this noise into the stereo-triangulated dense point cloud, we construct a high-fidelity dense ellipsoidal point map.

\begin{figure}[t]
\centering
\includegraphics[width=\linewidth]{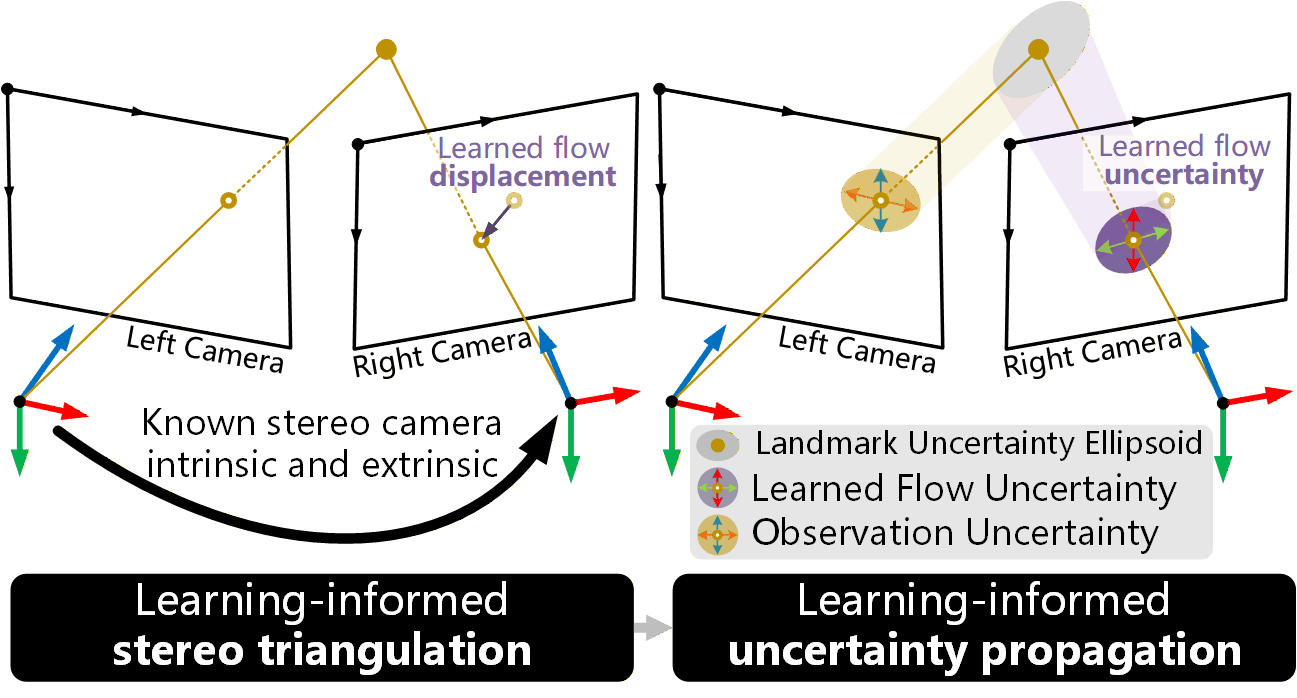}
\caption{Learning-informed stereo triangulation and uncertainty propagation}
\label{fig:ellipsoid}
\vspace{-3mm}
\end{figure}
First, we utilize the learned dense flow from the current stereo pair to triangulate 3D landmarks with respect to the left camera frame via the Direct Linear Transform. We then propagate the 2D covariance into a 3D point information matrix (Hessian) using the projection Jacobians (see Fig.~\ref{fig:ellipsoid}):
\begin{equation}
         \mathbf{H} =  \mathbf{J}_l^\intercal \boldsymbol{\Sigma}_{\text{obs}}^{-1} \mathbf{J}_l+ \mathbf{J}_r^\intercal \boldsymbol{\Sigma}_{\text{pro}}^{-1} \mathbf{J}_r
\end{equation}
where $\mathbf{J}_l$ and $\mathbf{J}_r$ are the projection Jacobians for the left and right views. To align with the global rendering pipeline, this information matrix is rotated into the world frame via $\mathbf{R}_{WC}$ and inverted to yield the final 3D landmark covariance:
\begin{equation}
   \boldsymbol{\Sigma}_{\mathbf{p}}= \left(\mathbf{R}_{WC}\;\mathbf{H}\;\mathbf{R}^{\intercal}_{WC}\right)^{-1}.
\end{equation}

\begin{figure}[t]
\centering
\includegraphics[width=0.99\linewidth]{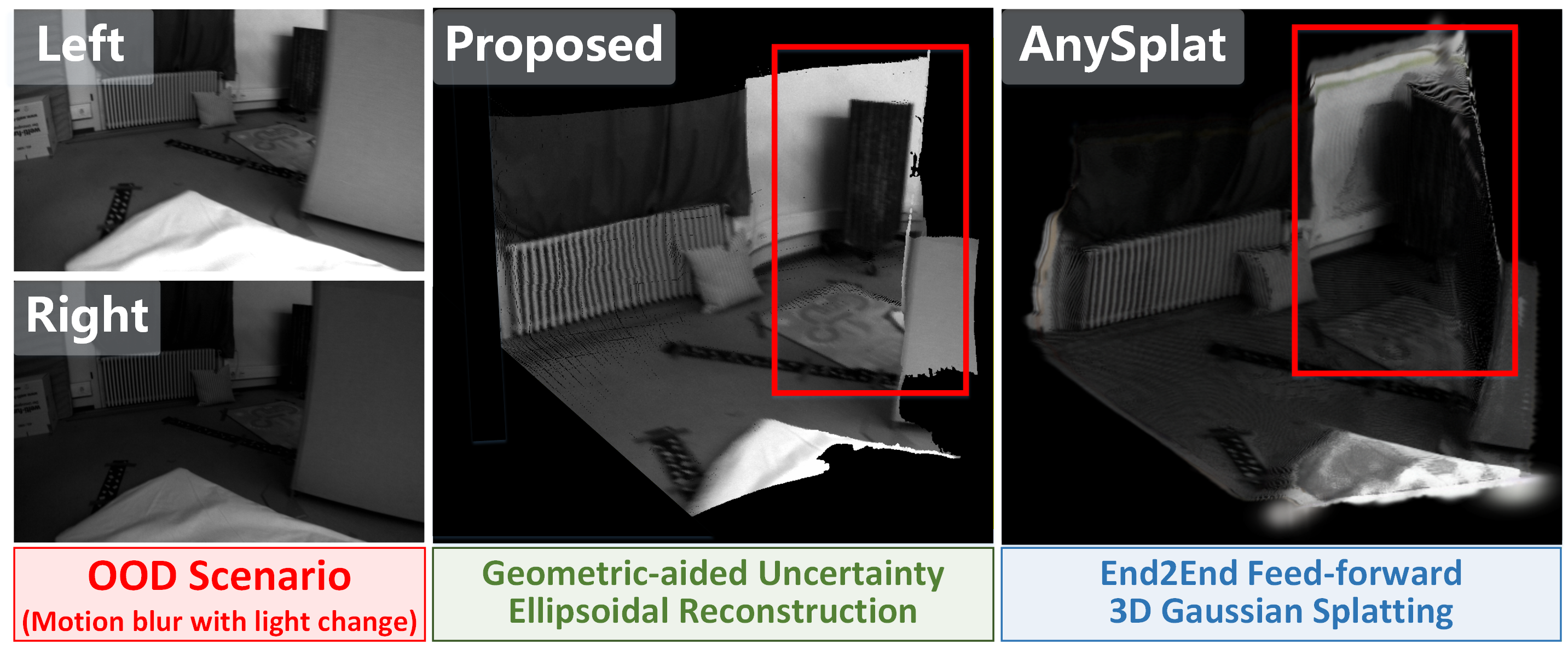}
\caption{Qualitative 3D Gaussian mapping results on an out-of-distribution (OOD) sequence. The proposed method uses uncertainty-aware stereo triangulation, while AnySplat\cite{RN1416} relies on geometry produced by VGGT. This result shows that the proposed uncertainty is not only useful for VIO state estimation, but also provides a reliable geometric prior for downstream 3D Gaussian mapping. Under OOD conditions, VGGT-based geometry may produce blurred or inconsistent Gaussian reconstruction, while the proposed uncertainty-aware mapping preserves sharper structures and more coherent appearance.}
\label{fig:3DGSresult}
\vspace{-3mm}
\end{figure}

The qualitative 3D Gaussian mapping is shown in Fig.~\ref{fig:3DGSresult}, where we compare the reconstruction fidelity of our proposed uncertainty-aware stereo triangulation against AnySplat\cite{RN1416} on a challenging OOD sequence.

\section{Experiments}
The experimental evaluation is designed to address three questions.
First, does learned dense correspondence improve VIO robustness under OOD conditions?
Second, are geometric verification and covariance modeling necessary for reliably incorporating learned correspondences into a VIO estimator?
Third, can the propagated uncertainty benefit both state estimation and downstream 3D Gaussian mapping?

\noindent \textbf{Dataset Selection.}
We evaluate on EuRoC \cite{RN314}, VIODE \cite{minoda2021viode}, and 4Seasons \cite{RN316} to cover progressively stronger distribution shifts, from indoor motion blur and illumination changes, to dynamic simulated scenes, and finally challenging real-world outdoor conditions involving traffic, day--night variation, reflections, fogging, and rain. Since the learned components are trained primarily on TartanAir \cite{wang2020tartanair}, all three datasets serve as unseen test domains without task-specific fine-tuning.

\noindent \textbf{Baseline Selection.}
We compare against three groups of methods selected to provide representative and reproducible coverage of major VO/VIO paradigms. We prioritize widely recognized methods with public implementations or sufficiently documented evaluation protocols. The first group contains classical geometric VO/VIO systems, including VINS-Mono \cite{RN91}, OKVIS2-VIO \cite{leutenegger2022okvis2}, DM-VIO \cite{RN455}, ORB-SLAM3 \cite{campos2021orb}, SchurVINS \cite{RN1291}, RD-VIO \cite{RN1279}, DynaVINS \cite{song2022dynavins}, and BASALT \cite{RN474}. The second group contains learning-enhanced geometric methods, including AirSLAM \cite{RN1293} and MAC-VO \cite{qiu2025mac}. The third group comprises learning-dominant methods, including DROID-VO \cite{teed2021droid}, DPVO \cite{teed2023deep}, TartanVO \cite{wang2021tartanvo}, and VGGT-SLAM \cite{maggio2026vggt}. DROID-VO, DPVO, and TartanVO are evaluated using models trained on TartanAir \cite{wang2020tartanair}. Together, these baselines represent three major design choices: purely geometric estimation, learned visual measurements with geometric optimization, and learning-dominant motion or geometry estimation. To focus on odometry performance and reduce the influence of global correction, loop closure is disabled whenever supported by the public implementation.

\noindent \textbf{Evaluation Protocol and Metric.}
The Root Mean Square Error (RMSE) of the Absolute Trajectory Error (ATE) is used as the main evaluation metric \cite{RN376}.
For monocular methods with scale ambiguity, Sim(3) alignment is applied following common VO evaluation practice, whereas SE(3) alignment is used for metric stereo-inertial systems. Each applicable method and ablation configuration is evaluated five times on every sequence, and the reported ATE values are averaged over the five runs. Overall, the evaluation comprises over 3,000 sequence-level runs, which required about four weeks of cumulative benchmark execution on our evaluation platform. Each method is executed in a separate Conda environment following recommended dependencies and configurations.

\begin{table*}[t]
\centering
\caption{Unified trajectory estimation results on EuRoC \cite{RN314}, VIODE \cite{minoda2021viode}, and 4Seasons \cite{RN316}.}
\vspace{-5pt}
\label{tab:all_results}
\setlength{\tabcolsep}{2pt}
\resizebox{\textwidth}{!}{
\begin{tabular}{l|ccccccccccc|cccccccccccc|ccccccc}


\toprule
\multirow{3}{*}[-0.5mm]{\fourdiagcell}
& \multicolumn{11}{c|}{\textbf{EuRoC}}
& \multicolumn{12}{c|}{\textbf{VIODE}}
& \multicolumn{7}{c}{\textbf{4Seasons}} \\
\cmidrule(lr){2-12}
\cmidrule(lr){13-24}
\cmidrule(lr){25-31}

& \texttt{MH01} & \texttt{MH02} & \texttt{MH03}
& \texttt{MH04} & \texttt{MH05}
& \texttt{V101} & \texttt{V102} & \texttt{V103}
& \texttt{V201} & \texttt{V202} & \texttt{V203}
& \texttt{cd-n} & \texttt{cd-l} & \texttt{cd-m} & \texttt{cd-h}
& \texttt{cn-n} & \texttt{cn-l} & \texttt{cn-m} & \texttt{cn-h}
& \texttt{pl-n} & \texttt{pl-l} & \texttt{pl-m} & \texttt{pl-h}
& \texttt{ot1} & \texttt{ot2} & \texttt{ot3} & \texttt{ot4}
& \texttt{cl1} & \texttt{cl2} & \texttt{cl3} \\

& 81 & 73 & 131 & 92 & 98
& 56 & 76 & 79 & 36 & 83 & 86
& 158 & 158 & 158 & 158
& 166 & 166 & 166 & 166
& 76 & 76 & 76 & 76
& 5.2k & 5.1k & 4.8k & 4.6k
& 11k & 10.3k & 10.3k \\

\midrule

DM-VIO\texttt{-M-I}
& 0.07 & \cellcolor{yellow!30}0.04 & 0.1 & \cellcolor{yellow!30}0.1 & \cellcolor{red!30}0.1 & \cellcolor{red!30}0.05 & 0.05 & 0.07 & \cellcolor{green!30}0.03 & \cellcolor{red!30}0.05 & 0.11
& 0.21 & \cellcolor{green!30}0.05 & \cellcolor{red!30}0.14 & \cellcolor{red!30}0.14 & 0.5 & \xmark   & \xmark   & \xmark   & 0.24 & 0.21 & \cellcolor{red!30}0.26 & 0.18
& \cellcolor{red!30}38.4 & \cellcolor{yellow!30}22.4 & \cellcolor{yellow!30}31.0 & \cellcolor{red!30}74.2 & \cellcolor{yellow!30}248.5 & \xmark  & \xmark   \\

VINS-Mono\texttt{-M-I}
& 0.18 & 0.09 & 0.17 & 0.21 & 0.25 & 0.06 & 0.09 & 0.18 & 0.06 & 0.11 & 0.26
& 0.17 & 0.14 & 0.59 & 3.21 & 0.47 & 0.83 & 0.62 & 0.98 & 3.17 & 0.17 & 1.69 & 2.89
& 50.8 & \cellcolor{red!30}32.6 & \xmark & \xmark & \xmark & \xmark & \xmark  \\

RD-VIO\texttt{-M-I}
& 0.11 & 0.12 & 0.14 & 0.25 & 0.27 & 0.06 & 0.09 & 0.13 & 0.06 & 0.1 & 0.15
& 0.23 & 0.29 & 0.24 & 0.44 & 0.96 & \cellcolor{red!30}0.8 & 0.81 & 0.79 & 0.25 & 0.15 & 0.4 & 0.23
& 502.5 & 547.4 & 449.1 & 456.3 & \xmark & \xmark & \xmark    \\

OKVIS2-VIO\texttt{-S-I}
& \cellcolor{red!30}0.06 & \cellcolor{yellow!30}0.04 & 0.08 & 0.19 & 0.14 & \cellcolor{yellow!30}0.04 & \cellcolor{red!30}0.04 & \cellcolor{yellow!30}0.04 & \cellcolor{yellow!30}0.04 & \cellcolor{yellow!30}0.04 & \cellcolor{yellow!30}0.06
& 0.21 & 0.3 & 0.42 & 0.66 & \cellcolor{red!30}0.38 & 0.94 & 0.83 & \cellcolor{red!30}0.73 & 0.11 & \cellcolor{red!30}0.07 & \cellcolor{yellow!30}0.06 & \cellcolor{red!30}0.08
& 146.2 & 286.8 & 164.8 & \xmark & \xmark & 725.9 & \cellcolor{red!30}455.9 \\


BASALT\texttt{-S-I}
& 0.07 & \cellcolor{red!30}0.06 & \cellcolor{red!30}0.07 & \cellcolor{red!30}0.13 & 0.11 & \cellcolor{yellow!30}0.04 & 0.05 & 0.1 & \cellcolor{yellow!30}0.04 & \cellcolor{red!30}0.05 & 0.24
& 0.41 & 0.44 & \xmark   & \xmark   & 0.7 & 2.18 & 3.34 & 3.8 & 0.09 & 0.19 & \xmark   & \xmark
& 94.9 & 251.8 & 91.1 & 104.6 & \xmark & \cellcolor{red!30}726.8 & \cellcolor{yellow!30}439.4 \\

SchurVINS\texttt{-S-I}
& \cellcolor{yellow!30}0.05 & 0.08 & 0.09 & \cellcolor{red!30}0.13 & 0.13 & \cellcolor{yellow!30}0.04 & 0.05 & 0.08 & \cellcolor{red!30}0.05 & 0.08 & 0.08
& \cellcolor{red!30}0.11 & 0.14 & \cellcolor{green!30}0.07 & \cellcolor{green!30}0.09 & \xmark & \xmark & \cellcolor{red!30}0.3 & \cellcolor{yellow!30}0.32 & 0.11 & 0.13 & \xmark & \xmark
& \xmark & \xmark & \xmark & \xmark & \xmark   & \xmark & \xmark \\

DynaVINS\texttt{-S-I}
& 0.31 & 0.15 & 1.79 & 2.26 & \xmark & \xmark & 0.37 & \xmark & \xmark & \xmark & \xmark 
& 0.17 & 0.18 & \cellcolor{yellow!30}0.09 & 0.15 & \cellcolor{yellow!30}0.21 & \cellcolor{green!30}0.18 & \cellcolor{yellow!30}0.2 & \cellcolor{green!30}0.2 & \cellcolor{green!30}0.05 & \cellcolor{green!30}0.04 & \cellcolor{yellow!30}0.06 & \cellcolor{green!30}0.04
& \xmark & \xmark & \xmark & \xmark & \xmark & \xmark & \xmark   \\


ORB-SLAM3\texttt{-S-I}
& \cellcolor{green!30}0.04 & \cellcolor{green!30}0.03 & \cellcolor{green!30}0.04 & \cellcolor{green!30}0.05 &\cellcolor{yellow!30} 0.08 & \cellcolor{yellow!30}0.04 & \cellcolor{green!30}0.01 & \cellcolor{green!30}0.02 & \cellcolor{green!30}0.03 & \cellcolor{green!30}0.01 & \cellcolor{green!30}0.02   
& 0.31 & 0.41 & 0.3 & \xmark   & 4.15 & 0.94 & 2.2 & \xmark   & \cellcolor{red!30}0.07 & 0.09 & \xmark   & \xmark
&\cellcolor{yellow!30}23.12 & \xmark  & \xmark & \xmark &\xmark & \xmark & \xmark \\

$^{*}$DROID-VO\texttt{-M}
& 0.16 & 0.12 & 0.24 & 0.4 & 0.27 & 0.1 & 0.17 & 0.16 & 0.1 & 0.12 & 0.2
& \cellcolor{yellow!30}0.1  & \cellcolor{yellow!30}0.08 & 3.84 & 2.61 & 5.69  & 9.52  & 12.01 & 10.42 & 2.25 & 1.95 & 5.24 & 3.76
& 269.7 & 669.4 & 285.8 & 595 & \xmark & \xmark & \xmark \\

$^{*}$DPVO\texttt{-M}
& 0.09 & 0.06 & 0.16 & 0.14 & 0.11 & 0.05 & 0.14 & 0.09 & 0.06 & \cellcolor{red!30}0.05 & 0.21
& \cellcolor{green!30}0.05 & 0.15 & 3.55 & \xmark  & 1.93 & 4.45 & 3.38 & \xmark & 0.65 & 0.73 & \xmark & 3.43
& 60.6 & 71.6 & \cellcolor{red!30}38.2 & \cellcolor{yellow!30}44 & \xmark &\cellcolor{yellow!30} 486.5 & 558 \\

$^{*}$TartanVO\texttt{-M}
& 0.64 & 0.33 & 0.55 & 1.15 & 1.02 & 0.45 & 0.39 & 0.62 & 0.43 & 0.75 & 1.15
& 5.09 & 5.47 & 4.83 & 4.23 & 8.14 & 8.08 & 8.22 & 8.6 & 1.46 & 1.41 & 8.51 & 8.38
& 643.7 & 755.2 & 710.1 & 879.8 & \xmark & \xmark & \xmark \\

$^{*}$VGGT-SLAM\texttt{-M}
& 0.81 & 0.67 & 2.3 & 2.43 & 2.21 & 0.55 & 4.59 & 0.98 & 1.05 & 1.2 & 5.78
& 6.37 & 6.39 & 6.35 & 6.09 & 12.66 & 12.52 & 12.07 & 13.15 & 2.94 & 2.94 & 3.28 & 2.92
& \xmark & \xmark & \xmark & \xmark & \xmark & \xmark & \xmark \\

AirSLAM\texttt{-S-I}
& 0.07 & \cellcolor{red!30}0.06 & 0.11 & 0.17 & 0.13 & \cellcolor{green!30}0.03 & 0.13 & 0.24 & \cellcolor{yellow!30}0.04 & 0.08 & 0.17
& 1.43 & 1.66 & 9.68 & 6.46 & 3.86 & 3.09 & 15.37 & 9.13 & 1.07 & 3.69 & 6.12 & 7.28
& 547.2 & 545  & 472.3 & 438.4 & \xmark & \xmark & \xmark    \\

MAC-VO\texttt{-S}
& 0.57 & 0.54 & 2.5 & 1.97 & 1.52 & 0.46 & 0.94 & 1.05 & 0.38 & 0.9 & 1.29
& 0.68 & 2.96 & 6.28 & 11.16 & 2.32 & 3.83 & 6.19 & 6.61 & 0.34 & 0.61 & 5.79 & 5.89
& 183.3 & 305.9 & 304.2 & 436.1 & \xmark & \xmark & 908.3  \\

\textbf{Proposed}\texttt{-S-I}
& \cellcolor{yellow!30}0.05 & \cellcolor{green!30}0.03 & \cellcolor{yellow!30}0.05 & \cellcolor{green!30}0.05 & \cellcolor{green!30}0.06 & \cellcolor{yellow!30}0.04 & \cellcolor{yellow!30}0.03 & \cellcolor{red!30}0.05 & \cellcolor{yellow!30}0.04 & \cellcolor{yellow!30}0.04 & \cellcolor{red!30}0.07 
& 0.17 & \cellcolor{red!30}0.11 & 0.18 & \cellcolor{yellow!30}0.13 & \cellcolor{green!30}0.16 & \cellcolor{yellow!30}0.19 & \cellcolor{green!30}0.15 & \cellcolor{green!30}0.2 & \cellcolor{yellow!30}0.06 & \cellcolor{yellow!30}0.06 & \cellcolor{green!30}0.05 & \cellcolor{yellow!30}0.06
& \cellcolor{green!30}10.8 & \cellcolor{green!30}10.2 & \cellcolor{green!30}10.7 & \cellcolor{green!30}12.5 & \cellcolor{green!30}42.3 & \cellcolor{green!30}23.2 & \cellcolor{green!30}44.7 \\
\midrule


\end{tabular}
}
\vspace{-3mm}
\begin{flushleft}
\scriptsize
\texttt{cd}: \texttt{city\_day}, \texttt{cn}: \texttt{city\_night}, \texttt{pl}: \texttt{parking\_lot}, 
\texttt{ot}: \texttt{Old Town}, \texttt{cl}: \texttt{City Loop}. 
${}^{*}$ denotes $\mathrm{Sim}(3)$ alignment in evaluation to compensate for scale. 
\texttt{-M}: Monocular, \texttt{-S}: Stereo, \texttt{-I}: with IMU. 
``\xmark'' denotes failures.  All values are ATE RMSE (m) with performance ranked from 
    {\setlength{\fboxsep}{1pt}\colorbox{green!30}{best}} to 
{\setlength{\fboxsep}{1pt}\colorbox{yellow!30}{second}} to 
{\setlength{\fboxsep}{1pt}\colorbox{red!30}{third}}. The rest of the paper follows the same notations unless otherwise specified. 
\end{flushleft}
\vspace{-23pt}
\end{table*}

\noindent \textbf{Implementation Details.} The system uses KLT for temporal sparse tracking and SEA-RAFT for learned stereo correspondence generation. 
Dense optical flow is sampled at sparse feature locations to preserve the computational structure of a sparse VIO backend. 
The learned correspondences are verified using network-predicted uncertainty and stereo epipolar consistency. 
Accepted observations are inserted into the sliding-window VIO backend as covariance-weighted reprojection factors. 
The backend follows a tightly coupled stereo-inertial formulation with IMU preintegration, Cauchy robust loss, and first-estimate Jacobians for marginalization \cite{RN525}. 
For mapping experiments, the uncertainty-informed 2D visual covariance is propagated through stereo triangulation to initialize and weight 3D Gaussian primitives.

\noindent\textbf{Supplementary Video.} The supplementary video presents our high-level motivation, additional trajectory plots, and further visual analysis of issues with narration.

\subsection{Main Results under Distribution Shift}


Table~\ref{tab:all_results} summarizes the main trajectory results under standard benchmark and shifted outdoor conditions.
The comparison covers three representative method families: classical geometric VIO, learning-enhanced geometric methods such as AirSLAM \cite{RN1293} and MAC-VO \cite{qiu2025mac}, and learning-dominant methods such as DROID-VO \cite{teed2021droid}, DPVO \cite{teed2023deep}, TartanVO \cite{wang2021tartanvo}, and VGGT-SLAM \cite{maggio2026vggt}.
This design allows us to examine whether stronger or deeper learned components necessarily improve robustness under distribution shift.

On EuRoC \cite{RN314}, most mature geometric and learning-based methods achieve reasonable accuracy, since the dataset mainly contains indoor motion blur and illumination changes without strong scene-level distribution shift. 
However, the performance gap becomes clearer on 4Seasons \cite{RN316}, especially in the Old Town and City Loop sequences. 
These sequences contain large-scale outdoor motion, traffic occlusions, lens fogging, raindrops, wet-road reflections, and strong illumination changes. 
Under these shifted conditions, classical systems may lose reliable visual tracks, while learning-dominant methods can degrade when their learned geometric priors do not match the deployment distribution.

In contrast, the proposed method achieves the best overall accuracy and stability in the OOD case. Compared with AirSLAM \cite{RN1293}, it uses learning only for dense stereo correspondence and uncertainty estimation. On outdoor 4Seasons sequences, AirSLAM’s line features mainly lie on road markings and curbs, offering limited geometric benefit, while dynamic objects further degrade point-line associations (See Supp. Video). Unlike learning-heavy methods, our approach does not directly trust learned motion, depth, or structure. Instead, correspondences are geometrically verified and incorporated as uncertainty-weighted VIO measurements, suggesting that restricting learning to measurement generation improves OOD transferability while preserving explicit geometric-inertial estimation.

\subsection{Ablation Studies on Association and Verification}

Table~\ref{tab:ablation} reports two groups of ablations. 
First, KLT temporal tracking with SEA-RAFT stereo matching provides the best balance between temporal stability and learned correspondence robustness. 
Second, progressively adding uncertainty filtering, epipolar checking, and covariance-weighted VIO consistently reduces the average ATE RMSE. 
This confirms that the gain does not come from learned flow alone, but from verifying and uncertainty-weighting learned correspondences before using them as VIO measurements.

\begin{table*}[t]
\centering
\caption{Dataset-level comparison by visual-estimation scope and back-end architecture. Representative methods are included for detailed comparison.}
\vspace{-5pt}
\label{tab:fe_be_scope}

\tiny
\setlength{\tabcolsep}{3.0pt}
\renewcommand{\arraystretch}{1.0}

\resizebox{\textwidth}{!}{
\begin{tabular}{llccllllccccc}
\toprule

\multirow[c]{2}{*}{\textbf{Arch.}} &
\multirow[c]{2}{*}{\textbf{Method}} &
\multirow[c]{2}{*}{\textbf{Yr.}} &
\multirow[c]{2}{*}{\textbf{In.}} &
\multicolumn{2}{c}{\textbf{Front-end}} &
\multicolumn{2}{c}{\textbf{Back-end}} &
\textbf{Params.} &
\textbf{VRAM} &
\textbf{EuRoC} &
\textbf{VIODE} &
\textbf{4Seasons} \\

\cmidrule(lr){5-6}
\cmidrule(lr){7-8}

& & & &
\textbf{Type} &
\textbf{Learned} &
\textbf{Type} &
\textbf{Learned} &
\textbf{(M)} &
\textbf{(GB)} &
\textbf{ATE (SR)} &
\textbf{ATE (SR)} &
\textbf{ATE (SR)} \\

\midrule

C/C
& ORB-SLAM3 \cite{campos2021orb}
& 2021 & S-I
& C & None
& C & None
& 0 & 0
& \cellcolor{green!30}{0.034} (11/11)
& \cellcolor{red!30}1.06 (8/12)
& 23.1 (1/7) \\

C/C
& DM-VIO \cite{RN455}
& 2022 & M-I
& C & None
& C & None
& 0 & 0
&  \cellcolor{green!30}0.07 (11/11)
& \cellcolor{yellow!30}0.214 (9/12)
& \cellcolor{red!30}82.9 (5/7) \\

C/C
& SchurVINS \cite{RN1291}
& 2024 & S-I
& C & None
& C & None
& 0 & 0
&  \cellcolor{green!30}0.078 (11/11)
& \cellcolor{red!30}0.159 (8/12)
&  \xmark \quad (0/7) \\

\midrule

L/--
& TartanVO \cite{wang2021tartanvo}
& 2020 & M
& L & Motion/flow + pose reg.
& -- & None
& 47.3 & \cellcolor{green!30}{1.2}
&  \cellcolor{green!30}0.68 (11/11)
& \cellcolor{green!30} 6.04 (12/12)
& 747.2 (4/7) \\

L/H
& DROID-VO \cite{teed2021droid}
& 2021 & M
& L & Dense corr.
& H & Recurrent pose/depth upd.
& \cellcolor{yellow!30}4.1 & 28.3
&  \cellcolor{green!30}0.185 (11/11)
&  \cellcolor{green!30}4.79 (12/12)
& 455.0 (4/7) \\

L/H
& DPVO \cite{teed2023deep}
& 2023 & M
& L & Patch assoc.
& H & Recurrent patch updates
& \cellcolor{green!30}3.4 & \cellcolor{yellow!30}1.5
&  \cellcolor{green!30}0.105 (11/11)
& \cellcolor{yellow!30}2.04 (9/12)
& \cellcolor{yellow!30}209.8 (6/7) \\

\midrule

L/C
& AirSLAM \cite{RN1293}
& 2025 & S-I
& L & Point/line detect.+match.
& C & None
& 20.6 & 3.1
&  \cellcolor{green!30}0.112 (11/11)
&  \cellcolor{green!30}5.74 (12/12)
& 500.7 (4/7) \\

L/C
& MAC-VO \cite{qiu2025mac}
& 2025 & S
& L & Depth + flow + uncert.
& C & None
& 19.2 & 4.2
&  \cellcolor{green!30}1.1 (11/11)
&  \cellcolor{green!30}4.39 (12/12)
& \cellcolor{red!30}427.6 (5/7) \\

L/C
& VGGT-SLAM \cite{maggio2026vggt}
& 2025 & M
& L & Pose + geometry
& C & None
& 1256.5 & 6.6
&  \cellcolor{green!30}2.05 (11/11)
&  \cellcolor{green!30}7.31 (12/12)
& \xmark \quad (0/7) \\

\midrule

\textbf{H/C}
& \textbf{Proposed}
& \textbf{2026} & \textbf{S-I}
& \textbf{H} & \textbf{KLT + stereo flow/uncert.}
& \textbf{C} & \textbf{None}
& \cellcolor{red!30}{8.9} & \cellcolor{red!30}2.1
& \cellcolor{green!30}0.046 (11/11)
& \cellcolor{green!30}{0.127} (12/12)
& \cellcolor{green!30}{22.1} (7/7) \\

\bottomrule
\end{tabular}
}

\vspace{0.5mm}
\raggedright
\scriptsize
C: classical; L: learned; H: hybrid; M: monocular; S: stereo;
I: IMU; BA: bundle adjustment; assoc.: association; corr.:
correspondence; uncert.: uncertainty; upd.: updates.; --: no explicit back-end;
ATE is averaged over successful
sequences, with performance ranked by SR (Success Rate) in parentheses. Best markings are meant for SR. 
\vspace{-12pt}
\end{table*}

\begin{table}[t]
\centering
\caption{Ablation of correspondence design and verification pipeline.}
\vspace{-5pt}
\label{tab:ablation}

\resizebox{\linewidth}{!}{
\begin{tabular}{cccccc}
\toprule

\multicolumn{6}{c}{
\textbf{Temporal (T) / Stereo (S) Correspondence Design with Learned (SEA-RAFT) / Classical (KLT)}
} \\
\midrule
T-KLT & T-RAFT & S-KLT & S-RAFT
& EuRoC \cite{RN314}$\downarrow$ & 4Seasons \cite{RN316}$\downarrow$ \\
\midrule
\cmark & \xmark & \cmark & \xmark
&\cellcolor{yellow!30} 0.066 &\cellcolor{yellow!30} 48.3 \\

\xmark & \cmark & \xmark & \cmark
& 0.168 & 86.4 \\

\xmark & \cmark & \cmark & \xmark
&\cellcolor{red!30}0.184 &\cellcolor{red!30}98.1 \\

\cmark & \xmark & \xmark & \cmark
& \cellcolor{green!30}{0.046} & \cellcolor{green!30}{22.1} \\

\midrule

\multicolumn{6}{c}{
\textbf{Incremental Verification on Temporal-KLT + Stereo-SEA-RAFT}
} \\
\midrule
Setting & Uncertainty & Epipolar & Covariance
& EuRoC \cite{RN314}$\downarrow$ & 4Seasons \cite{RN316}$\downarrow$ \\
\midrule

Raw flow
& \xmark & \xmark & \xmark
&0.078 & 98.7 \\

Uncertainty filtering
& \cmark & \xmark & \xmark
& \cellcolor{red!30}0.068 & \cellcolor{red!30}45.1 \\

+ Epipolar verification
& \cmark & \cmark & \xmark
& \cellcolor{yellow!30}0.062 & \cellcolor{yellow!30}42.8 \\

\textbf{Full proposed}
& \cmark & \cmark & \cmark
& \cellcolor{green!30}{0.046} & \cellcolor{green!30}{22.1} \\

\midrule

\multicolumn{6}{c}{
\textbf{SEA-RAFT Iteration--Accuracy-Efficiency Trade-off}
} \\
\midrule
\multicolumn{2}{c}{Iterations}
& EuRoC \cite{RN314}$\downarrow$
& 4Seasons \cite{RN316}$\downarrow$
& \multicolumn{2}{c}{SEA-RAFT Latency (ms) $\downarrow$} \\
\midrule

\multicolumn{2}{c}{4}
& \cellcolor{yellow!30}0.047
& \cellcolor{red!30}{25.8}
& \multicolumn{2}{c}{\cellcolor{red!30}93.13$\pm$12.06} \\

\multicolumn{2}{c}{2}
& \cellcolor{red!30}{0.056}
& \cellcolor{yellow!30}{23.5}
& \multicolumn{2}{c}{\cellcolor{yellow!30}65.51$\pm$9.75} \\

\multicolumn{2}{c}{\textbf{1 (Proposed)}}
& \cellcolor{green!30}{0.046}
& \cellcolor{green!30}{22.1}
& \multicolumn{2}{c}{\cellcolor{green!30}{43.61$\pm$5.69}} \\

\bottomrule
\end{tabular}
}

\vspace{-2mm}
\end{table}

\subsection{Computational Efficiency and Resource Analysis}

To assess deployability, we evaluate computational overhead on a lower-end processing unit equipped with an Intel Core i5-13400F CPU and an NVIDIA RTX 3060 Ti GPU. Since EuRoC \cite{RN314} operates at 20~Hz, each frame has a nominal processing budget of 50~ms. We therefore employ parallel front-end and back-end threads.
The visual front-end combines FAST detection, KLT temporal tracking, and SEA-RAFT stereo matching. Features are distributed over $35\times35$-pixel grids using adaptive thresholds, with at most two features per grid. To reduce latency, we use the SEA-RAFT Small model trained from TartanAir \cite{wang2020tartanair} with a single run time update iteration. The back-end jointly optimizes IMU pre-integration and visual reprojection residuals using a Cauchy robust loss and manifold Levenberg--Marquardt optimization. The sliding window contains six recent frames and up to 19 keyframes; active landmarks are eliminated during optimization, while outdated frames are marginalized.

\subsection{Discussion under Distribution Shift}








Table~\ref{tab:all_results} shows that our method achieves the best accuracy and success rate on the shifted VIODE and 4Seasons benchmarks while still remaining competitive on EuRoC.
On EuRoC, most mature geometric and learning-based methods perform reasonably because the benchmark contains indoor motion blur, fast motion, and illumination changes.
However, the performance gap widens on VIODE \cite{minoda2021viode} and 4Seasons \cite{RN316}, where dynamic objects, traffic occlusions, lens contamination, wet-road reflections, and large-scale outdoor motion introduce stronger distribution shifts (see SR of VIODE in Tab. \ref{tab:fe_be_scope}).
Under these conditions, purely geometric methods may lose reliable visual tracks, while learning-dominant methods can degrade when their learned geometric priors mismatch the deployment scenes  (see SR of 4Seasons in Tab. \ref{tab:fe_be_scope}).
In contrast, our method maintains stable estimation by using learning only for visual correspondence proposal and uncertainty prediction, while leaving final state estimation to explicit geometric-inertial optimization.

The ablation study in Tab.~\ref{tab:ablation} further explains where the performance gain comes from. SEA-RAFT provides denser and more robust stereo correspondences than hand-crafted matching, but directly using learned optical flow is not sufficient for reliable VIO. When SEA-RAFT is used without geometric verification, incorrect or overconfident flow predictions can introduce inconsistent visual constraints into the estimator. By progressively adding uncertainty filtering, epipolar checking, temporal consistency checking, and covariance-weighted reprojection factors, the average ATE RMSE is consistently reduced. This confirms that the key contribution is not simply replacing KLT with SEA-RAFT, but converting learned correspondences into estimator-compatible visual measurements.

These results (Tab. \ref{tab:fe_be_scope} and Tab.~\ref{tab:ablation}) support the hypothesis: more learning is not necessarily the best choice for OOD-robust VIO. Methods with stronger learned motion, depth, or structure priors can compete on standard benchmarks, but their outputs are not always compatible with a probabilistic VIO estimator. Our method follows a more conservative learned-geometric design. Learning is used to propose dense stereo correspondences and estimate their reliability, while epipolar geometry, temporal reprojection consistency, and covariance-weighted optimization determine how much each learned observation should influence the trajectory. This design improves transferability because the estimator does not fully rely on learned scene priors under distribution shift.

\begin{table}[t]
\centering
\caption{Average module timing and memory footprint on EuRoC.}
\label{tab:system_evaluation}
\vspace{-5pt}
\setlength{\tabcolsep}{3pt}
\renewcommand{\arraystretch}{1.08}

\begin{adjustbox}{max width=\columnwidth}
\begin{tabular}{llcc}
\toprule
\textbf{Category}
& \textbf{Module / Resource}
& \textbf{Value}
& \textbf{Unit} \\
\midrule

\multirow{3}{*}{\makecell[l]{Front-end\\Thread}}
& FAST feature detection
& $1.12 \pm 1.18$
& ms \\
& KLT temporal tracking
& $1.89 \pm 0.45$
& ms \\
& SEA-RAFT stereo matching
& $43.61 \pm 5.69$
& ms \\

\midrule

\multirow{3}{*}{\makecell[l]{Back-end\\Thread}}
& Manifold optimization
& $35.35 \pm 10.56$
& ms \\
& Landmark marginalization
& $2.15 \pm 1.15$
& ms \\
& Frame marginalization
& $1.72 \pm 0.85$
& ms \\

\midrule

\multirow{2}{*}{Memory}
& CPU memory
& $4.13 \pm 0.46$
& GB \\
& GPU memory
& $2.21 \pm 0.16$
& GB \\

\bottomrule
\end{tabular}
\end{adjustbox}

\vspace{-3mm}
\end{table}
Table~\ref{tab:system_evaluation} summarizes the runtime and memory usage.
The parallel front-end and back-end achieve average latencies of
46.62~ms and 39.22~ms, respectively, supporting approximately
20~Hz processing. Reducing SEA-RAFT from four iterations to one decreases
its latency from $93.13\pm12.06$~ms to $43.61\pm5.69$~ms, while maintaining
comparable trajectory accuracy. The system uses $4.13\pm0.46$~GB of CPU
memory and $2.21\pm0.16$~GB of GPU memory.

Overall, experiments show that the method provides a balance between accuracy, robustness, and transferability. Compared with purely geometric systems, it benefits from learned dense correspondence under degraded visual conditions. Compared with learning-heavy odometry or visual geometry systems, it avoids trusting learned motion or structure predictions. As shown in Fig. \ref{fig:3DGSresult}, propagating the observation uncertainty through stereo triangulation provides more reliable initialization for the 3D Gaussian primitives, resulting in sharper and more coherent reconstruction. The results indicate that OOD-robust VIO benefits more from geometry-verified and uncertainty-weighted learned measurements than from increasing the amount of learning in the pipeline.

\section{Conclusion}
This work is motivated by a central question: \textit{Does modern VIO really require learning a larger fraction of the estimation pipeline when facing distribution shifts?} Through thousands of extensive OOD experiments across EuRoC \cite{RN314}, VIODE \cite{minoda2021viode}, and 4Seasons \cite{RN316} with original training done on TartanAir \cite{wang2020tartanair}, we empirically show that robust generalization does not require a heavy learning pipeline. Instead, restricting learning to stereo correspondence and uncertainty prediction, while retaining explicit temporal tracking, geometric verification, and visual--inertial state estimation, provides a more effective balance between robustness, transferability, and interpretability. Our results further demonstrate that learned correspondences should not be treated as deterministic measurements; their benefits are realized only when uncertainty-aware filtering, stereo epipolar verification, and covariance-weighted reprojection are jointly incorporated into the estimator. Nevertheless, the method may still \textbf{face challenges} under extreme visual degradation, substantial camera-model shifts, or dynamic scenes in which incorrect correspondences remain geometrically plausible. \textbf{Future work} will focus on improving uncertainty calibration and introducing motion-aware consistency checks to better reject such failure cases.




\bibliographystyle{IEEEtran}
\bibliography{cas-refs}

@inproceedings{RN1291,
  author = {Fan, Yunfei and Zhao, Tianyu and Wang, Guidong},
  title = {{SchurVINS}: Schur Complement-Based Lightweight Visual Inertial Navigation System},
  booktitle = {CVPR},
  year = {2024}
}

@article{kerbl20233dgaussian,
  author  = {Bernhard Kerbl and Georgios Kopanas and Thomas Leimk{\"u}hler and George Drettakis},
  title   = {{3D} Gaussian Splatting for Real-Time Radiance Field Rendering},
  journal = {ACM Trans. Graph.},
  year    = {2023}
}

@article{RN1279,
  author = {Li, Jinyu and Pan, Xiaokun and Huang, Gan and Zhang, Ziyang and Wang, Nan and Bao, Hujun and Zhang, Guofeng},
  title = {{RD-VIO}: Robust Visual--Inertial Odometry for Mobile Augmented Reality in Dynamic Environments},
  journal = {IEEE Trans. Vis. Comput. Graph.},
  year = {2024}
}

@inproceedings{wang2025vggt,
  author    = {Jianyuan Wang and Minghao Chen and Nikita Karaev and Andrea Vedaldi and Christian Rupprecht and David Novotny},
  title     = {{VGGT}: Visual Geometry Grounded Transformer},
  booktitle = {CVPR},
  year      = {2025}
}

@book{RN517,
  author = {Barfoot, Timothy D.},
  title = {State Estimation for Robotics},
  publisher = {Cambridge Univ. Press},
  year = {2017}
}

@article{RN304,
  author = {Cadena, Cesar and Carlone, Luca and Carrillo, Henry and Latif, Yasir and Scaramuzza, Davide and Neira, Jos{\'e} and Reid, Ian and Leonard, John J.},
  title = {Past, Present, and Future of Simultaneous Localization and Mapping: Toward the Robust-Perception Age},
  journal = {TRO},
  year = {2016}
}

@article{RN303,
  author = {Fraundorfer, Friedrich and Scaramuzza, Davide},
  title = {Visual Odometry: Part I: The First 30 Years and Fundamentals},
  journal = {IEEE Robot. Autom. Mag.},
  year = {2011}
}

@article{RN302,
  author = {Fraundorfer, Friedrich and Scaramuzza, Davide},
  title = {Visual Odometry: Part II: Matching, Robustness, Optimization, and Applications},
  journal = {IEEE Robot. Autom. Mag.},
  year = {2012}
}

@article{maggio2026vggt,
  title={Vggt-slam: Dense rgb slam optimized on the sl (4) manifold},
  author={Maggio, Dominic and Lim, Hyungtae and Carlone, Luca},
  journal={NeurIPS},
  year={2026}
}

@inproceedings{RN298,
  author = {Triggs, Bill and McLauchlan, Philip F. and Hartley, Richard I. and Fitzgibbon, Andrew W.},
  title = {Bundle Adjustment: A Modern Synthesis},
  booktitle = {Proc. Int. Workshop Vis. Algorithms (IWVA)},
  year = {2000}
}

@article{RN400,
  author = {Forster, Christian and Carlone, Luca and Dellaert, Frank and Scaramuzza, Davide},
  title = {On-Manifold Preintegration for Real-Time Visual--Inertial Odometry},
  journal = {TRO},
  year = {2017}
}

@article{RN568,
  author = {Hesch, Joel A. and Kottas, Dimitrios G. and Bowman, Sean L. and Roumeliotis, Stergios I.},
  title = {Consistency Analysis and Improvement of Vision-Aided Inertial Navigation},
  journal = {TRO},
  year = {2014}
}

@inproceedings{RN469,
  author = {Koestler, Lukas and Yang, Nan and Zeller, Niclas and Cremers, Daniel},
  title = {{TANDEM}: Tracking and Dense Mapping in Real-Time Using Deep Multi-View Stereo},
  booktitle = {CoRL},
  year = {2022}
}

@article{RN309,
  author = {Leutenegger, Stefan and Lynen, Simon and Bosse, Michael and Siegwart, Roland and Furgale, Paul},
  title = {Keyframe-Based Visual--Inertial Odometry Using Nonlinear Optimization},
  journal = {Int. J. Robot. Res.},
  year = {2015}
}

@article{RN91,
  author = {Qin, Tong and Li, Peiliang and Shen, Shaojie},
  title = {{VINS-Mono}: A Robust and Versatile Monocular Visual--Inertial State Estimator},
  journal = {TRO},
  year = {2018}
}

@inproceedings{RN398,
  author = {Geneva, Patrick and Eckenhoff, Kevin and Lee, Woosik and Yang, Yulin and Huang, Guoquan},
  title = {{OpenVINS}: A Research Platform for Visual--Inertial Estimation},
  booktitle = {ICRA},
  year = {2020}
}

@article{RN474,
  author = {Usenko, Vladyslav and Demmel, Nikolaus and Schubert, David and St{\"u}ckler, J{\"o}rg and Cremers, Daniel},
  title = {Visual--Inertial Mapping with Non-Linear Factor Recovery},
  journal = {RAL},
  year = {2020}
}

@inproceedings{RN809,
  author = {Shi, Jianbo and Tomasi, Carlo},
  title = {Good Features to Track},
  booktitle = {CVPR},
  year = {1994}
}

@inproceedings{RN541,
  author = {Rublee, Ethan and Rabaud, Vincent and Konolige, Kurt and Bradski, Gary},
  title = {{ORB}: An Efficient Alternative to {SIFT} or {SURF}},
  booktitle = {Proc. IEEE Int. Conf. Comput. Vis. (ICCV)},
  year = {2011}
}

@book{RN586,
  author = {Thrun, Sebastian and Burgard, Wolfram and Fox, Dieter},
  title = {Probabilistic Robotics},
  publisher = {MIT Press},
  year = {2005}
}

@inproceedings{RN484,
  author = {Hsiung, Jerry and Hsiao, Ming and Westman, Eric and Valencia, Rafael and Kaess, Michael},
  title = {Information Sparsification in Visual--Inertial Odometry},
  booktitle = {IROS},
  year = {2018}
}

@article{RN455,
  author = {von Stumberg, Lukas and Cremers, Daniel},
  title = {{DM-VIO}: Delayed Marginalization Visual--Inertial Odometry},
  journal = {RAL},
  year = {2022}
}

@article{RN533,
  author = {Barfoot, Timothy D. and Furgale, Paul T.},
  title = {Associating Uncertainty with Three-Dimensional Poses for Use in Estimation Problems},
  journal = {TRO},
  year = {2014}
}

@article{RN1156,
  author = {Mangelson, Joshua G. and Ghaffari, Maani and Vasudevan, Ram and Eustice, Ryan M.},
  title = {Characterizing the Uncertainty of Jointly Distributed Poses in the Lie Algebra},
  journal = {TRO},
  year = {2020}
}

@inproceedings{RN376,
  author = {Zhang, Zichao and Scaramuzza, Davide},
  title = {A Tutorial on Quantitative Trajectory Evaluation for Visual--Inertial Odometry},
  booktitle = {IROS},
  year = {2018}
}

@inproceedings{RN316,
  author = {Wenzel, Patrick and Wang, Rui and Yang, Nan and Cheng, Qing and Khan, Qadeer and von Stumberg, Lukas and Zeller, Niclas and Cremers, Daniel},
  title = {{4Seasons}: A Cross-Season Dataset for Multi-Weather {SLAM} in Autonomous Driving},
  booktitle = {GCPR},
  year = {2020}
}

@article{RN314,
  author = {Burri, Michael and Nikolic, Janosch and Gohl, Pascal and Schneider, Thomas and Rehder, Joern and Omari, Sammy and Achtelik, Markus W. and Siegwart, Roland},
  title = {The {EuRoC} Micro Aerial Vehicle Datasets},
  journal = {Int. J. Robot. Res.},
  year = {2016}
}

@inproceedings{RN1243,
  author = {Zhao, Yipu and Vela, Patricio A.},
  title = {Good Feature Selection for Least Squares Pose Optimization in {VO/VSLAM}},
  booktitle = {IROS},
  year = {2018}
}

@article{RN1242,
  author = {Zhao, Yipu and Vela, Patricio A.},
  title = {Good Feature Matching: Toward Accurate, Robust {VO/VSLAM} with Low Latency},
  journal = {TRO},
  year = {2020}
}

@inproceedings{RN1011,
  author = {MacTavish, Kirk and Barfoot, Timothy D.},
  title = {At All Costs: A Comparison of Robust Cost Functions for Camera Correspondence Outliers},
  booktitle = {CRV},
  year = {2015}
}

@inproceedings{wang2020tartanair,
  title={Tartanair: A dataset to push the limits of visual slam},
  author={Wang, Wenshan and Zhu, Delong and Wang, Xiangwei and Hu, Yaoyu and Qiu, Yuheng and Wang, Chen and Hu, Yafei and Kapoor, Ashish and Scherer, Sebastian},
  booktitle={IROS},
  year={2020}
}

@article{RN1199,
  author = {Yang, Heng and Antonante, Pasquale and Tzoumas, Vasileios and Carlone, Luca},
  title = {Graduated Non-Convexity for Robust Spatial Perception: From Non-Minimal Solvers to Global Outlier Rejection},
  journal = {RAL},
  year = {2020}
}

@inproceedings{RN470,
  author = {Yang, Nan and von Stumberg, Lukas and Wang, Rui and Cremers, Daniel},
  title = {{D3VO}: Deep Depth, Deep Pose and Deep Uncertainty for Monocular Visual Odometry},
  booktitle = {CVPR},
  year = {2020}
}

@inproceedings{RN1252,
  author = {Mildenhall, Ben and Srinivasan, Pratul P. and Tancik, Matthew and Barron, Jonathan T. and Ramamoorthi, Ravi and Ng, Ren},
  title = {{NeRF}: Representing Scenes as Neural Radiance Fields for View Synthesis},
  booktitle = {ECCV},
  year = {2020}
}

@phdthesis{RN766,
  author = {Forster, Christian},
  title = {Visual--Inertial Odometry and Active Dense Reconstruction for Mobile Robots},
  school = {University of Zurich},
  year = {2016}
}

@article{RN531,
  author = {Schneider, Thomas and Dymczyk, Marcin and Fehr, Marius and Egger, Kevin and Lynen, Simon and Gilitschenski, Igor and Siegwart, Roland},
  title = {{Maplab}: An Open Framework for Research in Visual--Inertial Mapping and Localization},
  journal = {RAL},
  year = {2018}
}

@inproceedings{RN525,
  author = {Dong-Si, Tue-Cuong and Mourikis, Anastasios I.},
  title = {Motion Tracking with Fixed-Lag Smoothing: Algorithm and Consistency Analysis},
  booktitle = {ICRA},
  year = {2011}
}

@article{minoda2021viode,
  author = {Minoda, Koji and Schilling, Fabian and W{\"u}est, Valentin and Floreano, Dario and Yairi, Takehisa},
  title = {{VIODE}: A Simulated Dataset to Address the Challenges of Visual--Inertial Odometry in Dynamic Environments},
  journal = {RAL},
  year = {2021}
}

@article{campos2021orb,
  author = {Campos, Carlos and Elvira, Richard and Rodr{\'i}guez, Juan J. G{\'o}mez and Montiel, Jos{\'e} M. M. and Tard{\'o}s, Juan D.},
  title = {{ORB-SLAM3}: An Accurate Open-Source Library for Visual, Visual--Inertial, and Multimap {SLAM}},
  journal = {TRO},
  year = {2021}
}

@inproceedings{teed2020raft,
  author = {Teed, Zachary and Deng, Jia},
  title = {{RAFT}: Recurrent All-Pairs Field Transforms for Optical Flow},
  booktitle = {ECCV},
  year = {2020}
}

@inproceedings{wang2024sea,
  author = {Wang, Yihan and Lipson, Lahav and Deng, Jia},
  title = {{SEA-RAFT}: Simple, Efficient, Accurate {RAFT} for Optical Flow},
  booktitle = {ECCV},
  year = {2024}
}

@article{kang2025deepof,
  author = {Kang, Jeongmin},
  title = {{DeepOF-VIO}: A Filter-Based Visual--Inertial Odometry Using Deep Optical Flow},
  journal = {IEEE Sens. J.},
  year = {2026}
}

@inproceedings{teed2023deep,
  author = {Teed, Zachary and Lipson, Lahav and Deng, Jia},
  title = {Deep Patch Visual Odometry},
  booktitle = {NeurIPS},
  year = {2023}
}

@inproceedings{kang2024visual,
  author = {Kang, Jeong Min and Sjanic, Zoran and Hendeby, Gustaf},
  title = {Visual--Inertial Odometry Using Optical Flow from Deep Learning},
  booktitle = {Proc. Int. Conf. Inf. Fusion (FUSION)},
  year = {2024}
}

@inproceedings{teed2021droid,
  author = {Teed, Zachary and Deng, Jia},
  title = {{DROID-SLAM}: Deep Visual {SLAM} for Monocular, Stereo, and {RGB-D} Cameras},
  booktitle = {NeurIPS},
  year = {2021}
}

@inproceedings{wang2021tartanvo,
  author = {Wang, Wenshan and Hu, Yaoyu and Scherer, Sebastian},
  title = {{TartanVO}: A Generalizable Learning-Based {VO}},
  booktitle = {CoRL},
  year = {2021}
}

@inproceedings{murai2025mast3r,
  author = {Murai, Riku and Dexheimer, Eric and Davison, Andrew J.},
  title = {{MASt3R-SLAM}: Real-Time Dense {SLAM} with {3D} Reconstruction Priors},
  booktitle = {CVPR},
  year = {2025}
}

@article{leutenegger2022okvis2,
  title={Okvis2: Realtime scalable visual-inertial slam with loop closure},
  author={Leutenegger, Stefan},
  journal={arXiv preprint arXiv:2202.09199},
  year={2022}
}

@article{song2022dynavins,
  title={DynaVINS: A visual-inertial SLAM for dynamic environments},
  author={Song, Seungwon and Lim, Hyungtae and Lee, Alex Junho and Myung, Hyun},
  journal={RAL},
  year={2022}
}

@article{RN1293,
  author = {Xu, Kuan and Hao, Yuefan and Yuan, Shenghai and Wang, Chen and Xie, Lihua},
  title = {{AirSLAM}: An Efficient and Illumination-Robust Point-Line Visual {SLAM} System},
  journal = {TRO},
  year = {2025}
}

@inproceedings{qiu2025mac,
 title={MAC-VO: Metrics-Aware Covariance for Learning-Based Stereo Visual Odometry},
 author={Qiu, Yuheng and Chen, Yutian and Zhang, Zihao and Wang, Wenshan and Scherer, Sebastian},
 booktitle={ICRA},
 year={2025}
}

@article{RN1416,
   author = {Jiang, Lihan and Mao, Yucheng and Xu, Linning and Lu, Tao and Ren, Kerui and Jin, Yichen and Xu, Xudong and Yu, Mulin and Pang, Jiangmiao and Zhao, Feng and Lin, Dahua and Dai, Bo},
   title = {AnySplat: Feed-forward 3D Gaussian Splatting from Unconstrained Views},
   journal = {ACM Transactions on Graphics},
   year = {2025}
}

@IEEEtranBSTCTL{IEEEexample:BSTcontrol,
  CTLuse_article_number     = "yes",
  CTLuse_paper              = "yes",
  CTLuse_forced_etal        = "yes",
  CTLmax_names_forced_etal  = "1",
  CTLnames_show_etal        = "1",
  CTLuse_alt_spacing        = "yes",
  CTLalt_stretch_factor     = "4",
  CTLdash_repeated_names    = "no",
  CTLname_format_string     = "{f.~}{vv~}{ll}{, jj}",
  CTLname_latex_cmd         = ""
}


 




\vfill

\end{document}